\documentclass[final]{cvpr}

\usepackage{times}
\usepackage{epsfig}
\usepackage{graphicx}
\usepackage{amsmath}
\usepackage{amssymb}

\usepackage[pagebackref=true,breaklinks=true,colorlinks,bookmarks=false]{hyperref}

\def\cvprPaperID{****} 
\def\confYear{CVPR 2021}

\begin{document}

\title{Team Mia at TextVQA Challenge 2021: Vision-and-Language Representation Learning with Pre-trained Sequence-to-Sequence Model}

\author{Yixuan Qiao\textsuperscript{1}, Hao Chen\textsuperscript{1}, Jun Wang\textsuperscript{1,2}, Shanshan Zhao\textsuperscript{2}, Yihao Chen\textsuperscript{2}, Xianbin Ye\textsuperscript{3}, Ziliang Li\textsuperscript{4}, \\
Xianbiao Qi\textsuperscript{2}, Peng Gao\textsuperscript{1}, Guotong Xie\textsuperscript{1,5,6}\\
\textsuperscript{1} SFE Deeplearning Platform, Ping An Health Technology, Beijing, China.\\
\textsuperscript{2} Peking University, Beijing, China.\\
\textsuperscript{3} Visual Computing Group, Ping An Property \& Casualty Insurance Company, Shenzhen, China.\\
\textsuperscript{4} Jinan University, Guangzhou, China. 
\textsuperscript{5} Central University of Finance and Economics, Beijing, China.\\
\textsuperscript{6} Ping An Health Cloud Company Limited., Shenzhen, China.\\
\textsuperscript{7} Ping An International Smart City Technology Co., Ltd., Shenzhen, China.\\
{\tt\small \{qiaoyixuan528, chenhao305, zhaoshanshan233, wangjun916, chenyihao291, qixianbiao321\}@pingan.com.cn}\\
{\tt\small \{gaopeng712, xieguotong\}@pingan.com.cn}, 
{\tt\small \ yexianbin@stu2019.jnu.edu.cn, 2020210983@cufe.edu.cn}
}

\maketitle

\begin{abstract}
  TextVQA requires models to read and reason about text in images to answer questions about them. Specifically, models need to incorporate a new modality of text present in the images and reason over it to answer TextVQA questions.
  In this challenge, we use generative model T5 for TextVQA task. Based on pre-trained checkpoint T5-3B from HuggingFace repository, two other pre-training tasks including masked language modeling(MLM) and relative position prediction(RPP) are designed to better align object feature and scene text. In the stage of pre-training, encoder is dedicate to handle the fusion among multiple modalities: question text, object text labels, scene text labels, object visual features, scene visual features. After that decoder generates the text 
  sequence step-by-step, cross entropy loss is required by default. We use a large-scale scene text dataset in pre-training and then fine-tune the T5-3B with the TextVQA dataset only.  
\end{abstract}

\begin{figure*}[t]
\centering
\includegraphics[width=0.8\linewidth]
{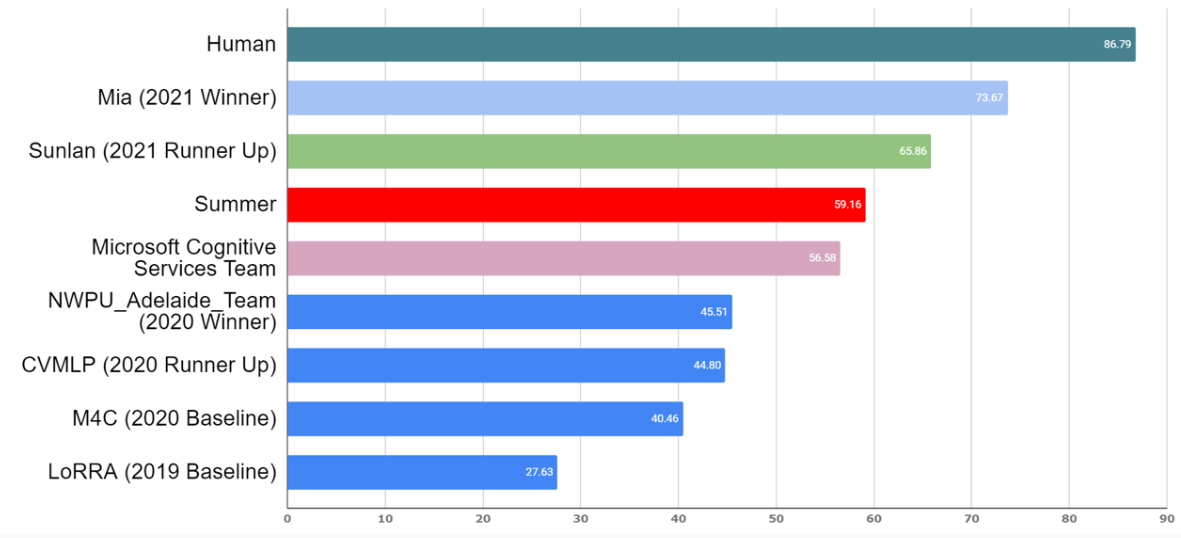}
\caption{Comparison of our solution and other teams from TextVQA 2021 and TextVQA 2020.}
\label{fig:leaderboard}
\end{figure*}

\section{Pre-training Stage}
\subsection{Pre-training Dataset}
Inspired by the promising results of TAP (Text-Aware Pre-training) ~\cite{yang2020tap}, further increasing the size of model and data are expected to have better performance. We built a dataset contains about 9 million scene text images. Part of it by using the similar methods as TAP, part of it by generating synthetic text images as described in ~\cite{gupta2016synthetic}. Meantime, we have corresponding relationships between object features and scene features for both parts.


\subsection{Pre-training Strategy}
First of all, for each scene image, we have the following features.

\begin{itemize}
    \item image texts, object labels, scene texts
    \item object visual features, scene visual features
\end{itemize}

Object labels and features are extracted from VinVL(Revisiting Visual Representations in Vision-Language Models) ~\cite{zhang2021vinvl}, which can generate representations of a richer collection of visual objects and concepts. Scene text and scene visual features are mainly coming from public Microsoft OCR API. 

For each scene visual features, scene text pair in images, we use the former with all other features as input, scene text as target to construct training samples. This design comes from the fact that half of the answers in the Text-VQA dataset came from scene text which is necessary to equip the model with the ability to generate scene text. For MLM on the extended text input, we randomly mask each text token with a probability of 15\%, the details are the same as BERT.
The objective of RPP is to predict the relative position between the scene visual feature and each object visual feature.

Additionally, to enable large-scale training, we adopt the adversarial training strategy combined with KL-divergence-based regularization to
promote higher invariance in the embedding space. Specifically, we propose to perform adversarial training in the embedding space of each modality, rather than adding adversarial perturbations on image pixels and textual tokens. Finally, together with the cross-entropy loss of decoder as final objective to be optimized.

\section{Fine-tuning stage }
In fine-tuning, the model step-by-step predicts
the answer using its decoding module, and is trained
with the answer classification loss and the adversarial training strategy above in each step.

\begin{table}[t!]
	\centering
	\resizebox{1.0\columnwidth}{!}{%
    \begin{tabular}{|c|c|}
    \hline
    \textbf{Metric}                               & \textbf{TextVQA} \\ \hline
    Number of images                              & 28k              \\ \hline
    Images with text                              & 100\%             \\ \hline
    Number of questions                           & 45k              \\ \hline
    Number of questions with answer in OCR tokens & 39\% (18k)        \\ \hline
    Number of questions with spatial words        & 14\%              \\ \hline
    \end{tabular}
}
\vspace{0.3cm}
\caption{Details of the TextVQA dataset in Fine-tuning stage. }

\end{table}


\section{Post-processing Module}
After we got the answer sequence through fine-tuning stage, we use a open source fuzzy string matching method fuzzywuzzy which uses Levenshtein Distance to calculate the differences between sequences to make the final correction.

\section{Results and Conclusions}

TextVQA requires models to read and reason about text in an image to answer questions based on them. In order to perform well on this task, models need to first detect and read text in the images. Models then need to reason about this to answer the question. There are 115 and 65 submissions in validation and test phases from the 33 teams participating in TextVQA 2021 challenge. The final comparison between our solution and other results from TextVQA challenges is shown in Figure \ref{fig:leaderboard}

To summarize, the experiments from this challenge present the following key suggestions: 
\begin{itemize}
\item Pretraining with image-text matching contrastive objectives on huge data leads to significant gains.
\item Developing hard-negatives for such objectives also improve performance.
\item There is a scope of improvement in multi-word answers.
\item Using a pretrained NLP model as a base helps.
\item Improved object and scene-text features also lead to some gains.
\end{itemize}

\bibliographystyle{ieee_fullname}
\bibliography{egbib}

\end{document}